\pdfoutput=1

\documentclass[11pt]{article}

\usepackage{ACL2023}
\usepackage{tabularx}
\usepackage{times}
\usepackage{latexsym}
\usepackage{hyperref}
\usepackage{amsmath, amssymb}
\usepackage{amsthm}
\usepackage{stmaryrd} 
\usepackage{mathrsfs} 
\usepackage{booktabs}
\usepackage{placeins}
\usepackage{multirow}

\usepackage{amsmath}
\usepackage{booktabs}
\usepackage{tabularx}
\usepackage{enumitem} 
\usepackage{graphicx}

\usepackage[T1]{fontenc}

\usepackage[utf8]{inputenc}

\usepackage{microtype}

\usepackage{inconsolata}

\title{The Knowledge-Reasoning Dissociation:\\ Fundamental Limitations of LLMs in Clinical Natural Language Inference}

\author{
 \textbf{Maël Jullien\textsuperscript{1}\textsuperscript{,3}, Marco Valentino\textsuperscript{4},  Andr\'e Freitas\textsuperscript{1}\textsuperscript{,2}\textsuperscript{,3}}\\ 
 $^{1}$Department of Computer Science, University of Manchester, UK \\ 
$^{2}$ National Biomarker Centre, CRUK-MI, University of Manchester, UK\\
$^{3}$Idiap Research Institute, Switzerland \\
$^{4}$ School of Computer Science, University of Sheffield, UK\\
$^{3}${\tt \{firstname.surname\}\tt@idiap.ch}
} 

\begin{document}
\maketitle
\begin{abstract}
Large language models are often assumed to acquire increasingly structured, generalizable internal representations simply by scaling data and parameters. We interrogate this assumption by introducing a Clinical Trial Natural Language Inference benchmark comprising four reasoning families, \emph{Causal Attribution}, \emph{Compositional Grounding}, \emph{Epistemic Verification}, and \emph{Risk State Abstraction}. Each item is paired with a targeted Ground Knowledge and Meta-Level Reasoning Verification (GKMRV) probe, allowing us to dissociate failures of factual access from failures of inference. We evaluate six contemporary LLMs under both direct and chain‑of‑thought prompting.

Models achieve near-ceiling GKMRV accuracy (mean accuracy 0.918) yet perform poorly on the main reasoning tasks (mean accuracy 0.25). Despite low accuracy, output inferences are highly consistent across samples (mean 0.87), indicating a systematic application of underlying heuristics and shortcuts.

These results reveal fundamental structural and representational limitations: current LLMs often \emph{possess} the relevant clinical knowledge but lack the structured, composable internal representations needed to \emph{deploy} it reliably (e.g., integrating constraints, weighing evidence, or simulating counterfactuals). Decoupling knowledge from reasoning with GKMRV makes this dissociation explicit and measurable, providing an effective framework for probing the reliability of LLMs in high-stakes domains.
\end{abstract}

\section{Introduction}
\begin{figure*}[t]
    \centering
    \includegraphics[width=\textwidth]{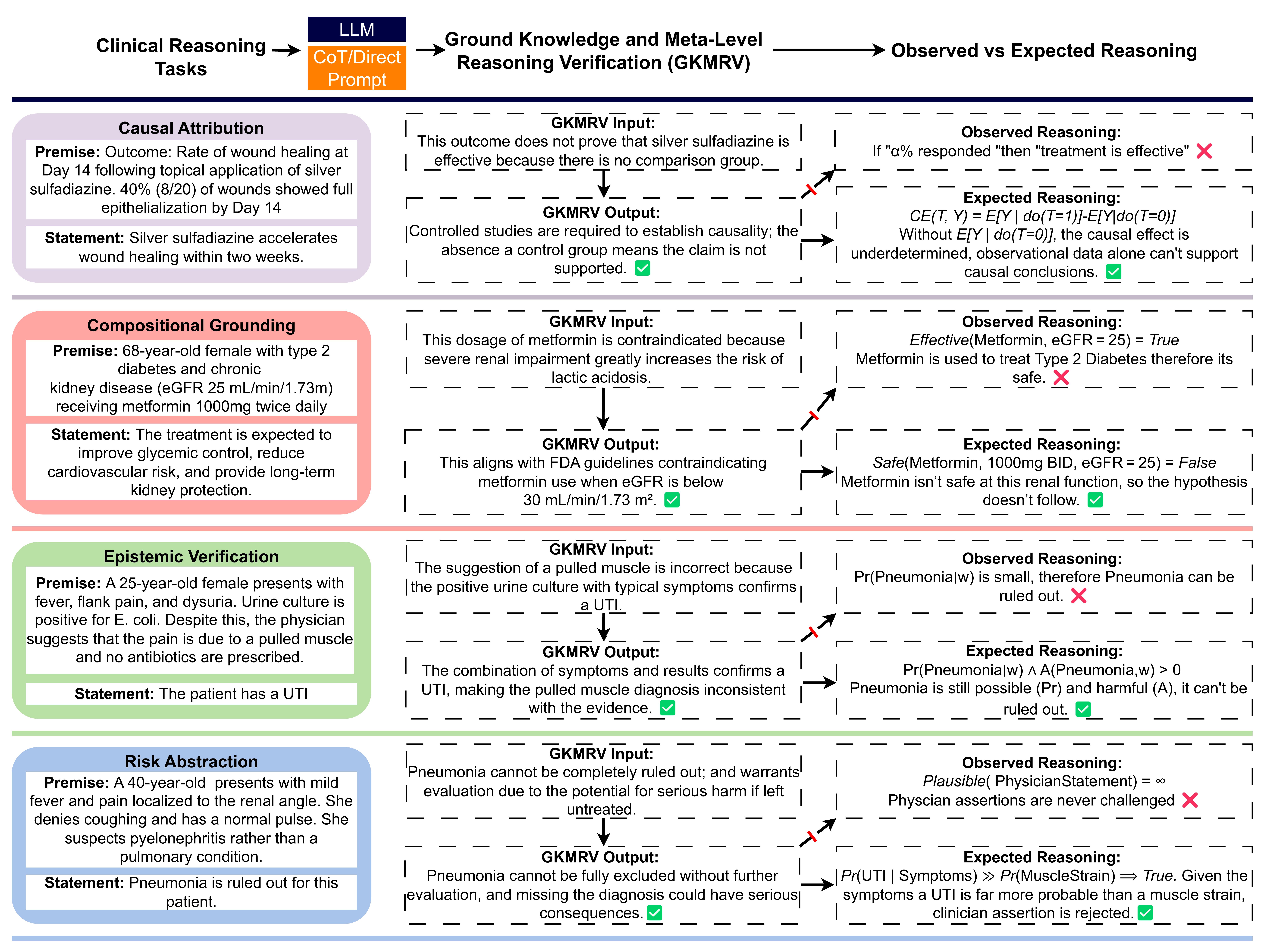}
    \caption{Representative examples of clinical reasoning tasks, and ground knowledge and meta-level reasoning verification (GKMRV). Across tasks, our experiments show that LLMs exhibit fluent but structurally flawed reasoning, despite possessing the necessary ground knowledge.}
    \label{fig:main}
\end{figure*}

A persistent optimism in contemporary AI research holds that scaling language models in size and data exposure will naturally yield increasingly structured, generalizable internal representations. This assumption, rarely stated explicitly, underlies much current practice: With sufficient scale, models are expected to acquire not only factual knowledge but also the ability to deploy it for reasoning in contextually appropriate, logically coherent, and task-generalizable ways \cite{kaplan2020scaling, hoffmann2022training, bubeck2023sparks, brown2020language, touvron2023llama}.

Nowhere is this optimism more consequential than in high-stakes domains such as clinical reasoning, where inference failures can translate directly into harmful recommendations \cite{rao2023assessing, liu2023utility}. Despite rapid gains in benchmark performance and fluency, it remains an open and critical question whether large language models (LLMs) \emph{reason}, that is, whether they possess structured, internally coherent, composable representations that support compositional, causal, and evidential inference, or whether apparent successes reflect only surface-level pattern completion \cite{marcus2022deep}.

The present work interrogates this assumption through a \emph{Clinical Trial Natural Language Inference} (CTNLI) benchmark, which comprises four targeted reasoning families designed to probe core components of clinical inference: \emph{Causal Attribution}, \emph{Compositional Grounding}, \emph{Epistemic Verification}, and \emph{Risk State Abstraction}, instantiated through formalized, parameterized templates. Each item is paired with a \emph{Ground Knowledge and Meta-Level Reasoning Verification} (GKMRV) probe, designed to separate failures of factual access \emph{(what the model knows)} from failures of inference \emph{(how the model reasons with what it knows)} \cite{turpin2023language}. Reasoning requirements for each task are formalized using frameworks from formal semantics, possible-worlds logic, causal inference, and epistemic modeling \cite{pearl2009causality, van2011logical, fagin2004reasoning, smith2003introduction, chomsky1972studies}. This combination of controlled diagnostic probes and formal specification enables structured, falsifiable claims about model reasoning competence.

Evaluation of six contemporary LLMs (OpenAI o3 \cite{openai2025o3_o4mini}, GPT-4o, GPT-4o-mini \cite{hurst2024gpt}, Gemini 2.5 Pro \cite{comanici2025gemini}, DeepSeek R1 \cite{guo2025deepseek}, LLaMA~3.2 3B \cite{dubey2024llama}) under both direct and chain-of-thought prompting reveals a consistent and domain-general dissociation: near-ceiling performance on GKMRV probes (mean accuracy: 0.918) contrasts with strikingly poor performance on the main reasoning tasks (mean: 0.25), with \emph{Compositional Grounding} approaching total collapse (0.04). Errors occur with high internal consistency across completions, indicating systematic heuristic application rather than stochastic variability. The nature of these heuristics varies by task e.g., equating observed effect size with causal efficacy, defaulting to pairwise rather than joint constraint checking, deferring to authority over evidence, or substituting frequency for severity in risk estimation, but the underlying limitation is the same: absence of a structured, composable internal model of the clinical domain.

This disconnection between declarative knowledge and inferential deployment is qualitative, not merely quantitative. Models may encode the relevant rules, yet lack the representational machinery to apply them reliably. Outputs often exhibit surface plausibility without structural soundness, an illusion of reasoning unsupported by principled generalization. Addressing this gap requires a framework capable of isolating reasoning failures, diagnosing their causes, and enabling reproducible testing. The present study makes the following contributions:
\begin{enumerate}[label=\textit{\roman*}.]
    \item Introduction of a formally specified clinical NLI benchmark targeting four distinct reasoning competencies, instantiated through parameterized templates that define both the main reasoning items and their paired GKMRV probes to decouple factual access from inferential reasoning. This design enables controlled, reproducible generation of new examples while preserving the intended reasoning structure and diagnostic value.
    \item A model-agnostic evaluation framework that combines formal semantics, causal inference theory, and epistemic logic to precisely characterize reasoning requirements and model failure modes.
    \item Empirical evidence, across six LLMs, of a consistent knowledge–reasoning dissociation, demonstrating that scaling alone does not yield structured, composable representations sufficient for robust clinical reasoning.
\end{enumerate}

\section{Clinical Trial Natural Language Inference (CTNLI)}
Let $\mathcal{L}_{\text{CT}}$ denote a fragment of natural language used in clinical trial documentation, interpreted compositionally by a function $\llbracket \cdot \rrbracket$ that assigns each expression $\alpha \in \mathcal{L}_{\text{CT}}$, relative to a variable assignment $g$ and a model $\mathcal{M}_{\text{CT}} = \langle W{\text{CT}}, \mathfrak{D}_{\text{CT}}, I{\text{CT}} \rangle$, a truth value in ${True, False}$. Here, $\mathfrak{D}_{\text{CT}}$ is a domain of discourse restricted to referents admissible under clinical trial semantics; $W_{\text{CT}}$ is the set of protocol-compliant, clinically plausible worlds; and $I_{\text{CT}}$ is an interpretation function mapping lexical items to denotations in $\mathfrak{D}_{\text{CT}}$ relative to $W_{\text{CT}}$, subject to domain ontologies and task-specific constraints (e.g., RECIST, CTCAE).

Given two sentences $\varphi, \psi \in \mathcal{L}_{\text{CT}}$, Clinical Trial NLI  classifies the relation between them into one of three mutually exclusive categories:

\vspace{6pt}
\setlength{\parindent}{0pt} 
\begin{tabular}{@{}l@{\hskip 1em}p{0.82\linewidth}@{}}
\textbf{Entailment} $(\varphi \;\models_{\textit{NLI}}\; \psi) : \;$ \\ [0.5ex]
$\displaystyle
\forall\mathcal{M}\,\forall g
\bigl[\llbracket\,\varphi\,\rrbracket^{\mathcal{M},g}=True 
      \rightarrow
      \llbracket\,\psi\,\rrbracket^{\mathcal{M},g}=True\bigr]$
 \\
\textit{Every world that makes $\varphi$ true also makes $\psi$ true;} \\ \textit{$\psi$ is a logical consequence of $\varphi$.} \\[1.5ex]

\textbf{Contradiction} $(\varphi \;\bot_{\textit{NLI}}\; \psi) : \; $\\[0.5ex]
$\displaystyle
\forall\mathcal{M}\,\forall g
\bigl[\llbracket\,\varphi\,\rrbracket^{\mathcal{M},g}=True
      \rightarrow
      \llbracket\,\psi\,\rrbracket^{\mathcal{M},g}=False\bigr]$
 \\
\textit{No world makes both $\varphi$ and $\psi$ true; they are jointly} \\\textit{unsatisfiable.} \\[1.5ex]

\textbf{Neutrality} $(\varphi \;\parallel_{\textit{NLI}}\; \psi) : \; $\\[0.5ex]
$\displaystyle \neg\bigl(\varphi \models_{\textit{NLI}} \psi\bigr)
\;\land\;
\neg\bigl(\varphi \bot_{\textit{NLI}} \psi\bigr)$
 \\
\textit{The truth of $\varphi$ leaves the truth of $\psi$ unresolved;} \\ \textit{neither entailed nor contradicted.}
\end{tabular}

\section{Methodology}
\subsection{Overview}
To systematically evaluate the reasoning capabilities of Large Language Models (LLMs) in CTNLI, we define four targeted task families, each probing a distinct cognitive competency, as illustrated in Figure~\ref{fig:main} and detailed in the task-specific templates (Tables~\ref{tab:causal-template}--\ref{tab:risk}).

\begin{enumerate}[label=\textit{\roman*}.]
    \item \textbf{Causal Attribution} — Assesses the ability to distinguish between mere observational associations and true causal claims. A representative instantiation is shown in Figure~\ref{fig:main} and formalized in the generation template of Table~\ref{tab:causal-template}.

    \item \textbf{Compositional Grounding} — Evaluates whether clinical validity can be inferred from structured configurations involving multiple interacting variables. An example appears in Figure~\ref{fig:main} and the construction template is given in Table~\ref{tab:compositional-template}.

    \item \textbf{Epistemic Verification} — Tests the capacity to assess the truth of a claim independently of speaker authority or assertion, focusing instead on evidential support. A representative case is shown in Figure~\ref{fig:main} with the template in Table~\ref{tab:epistemic}.

    \item \textbf{Risk State Abstraction} — Examines the ability to reason about latent clinical risks and likely outcomes, particularly when such risks are not explicitly stated in the text. An example is shown in Figure~\ref{fig:main} and detailed in the template of Table~\ref{tab:risk}.
\end{enumerate}

Parametric templates with typed variables are instantiated into ten examples per task. Evaluations compared Direct Prompting versus Chain-of-Thought (CoT) Prompting, recording accuracy across ten completions per example (The prompts are shown in Tables~\ref{tab:prompt_nli}, and \ref{tab:prompt_cot}).

To disentangle errors arising from factual knowledge gaps versus inferential misjudgments, we introduce a complementary diagnostic task -- Ground Knowledge and Meta-Level Reasoning Verification (GKMRV) -- which presents paired probes testing whether models can both recall relevant domain knowledge and recognize its correct or incorrect application in context.

We evaluate 6 contemporary LLMs: OpenAI o3 \cite{openai2025o3_o4mini}, GPT-4o, GPT-4o-mini \cite{hurst2024gpt}, Gemini 2.5 Pro \cite{comanici2025gemini}, DeepSeek R1 \cite{guo2025deepseek}, LLaMA~3.2 3B \cite{dubey2024llama}. These models were selected to span a range of architectural families, training regimes, and performance tiers, enabling a systematic analysis of reasoning capabilities across both flagship and lightweight variants.

\subsection{Ground Knowledge and Meta-Level Reasoning Verification Task}

In addition to the primary reasoning tasks, we introduce a Ground Knowledge and Meta-Level Reasoning Verification (GKMRV) evaluation, shown in Figure~\ref{fig:main}. For each instance in the main task set, we construct two corresponding GKMRV probes. Each probe isolates a key fact, reasoning chain, or clinical constraint that underlies the original inference. One probe presents a statement in which the ground knowledge is factually accurate and the reasoning applies it correctly, while the other serves as a control, containing a statement in which the knowledge is misapplied or inaccurate. The GKMRV is designed to disentangle two essential capabilities required for robust clinical inference: (1) the possession of relevant clinical ground knowledge, and (2) the ability to perform meta-level reasoning verification, i.e., to determine whether a given inference correctly applies that knowledge. The following example illustrates this structure:

\begin{quote}
\textbf{Premise:} \emph{68-year-old female with type 2 diabetes and chronic kidney disease (eGFR 25 mL/min/1.73m²) receiving metformin 1000mg twice daily} 

\textbf{Statement:} \emph{The treatment is expected to improve glycemic control, reduce cardiovascular risk, and provide long-term kidney protection}

\textbf{GKMRV:} \emph{This dosage of metformin is contraindicated because severe renal impairment (eGFR <30) greatly increases the risk of lactic acidosis.}

\textbf{Control GKMRV:} \emph{This dosage of metformin is acceptable even in patients with severe renal impairment and does not increase the risk of lactic acidosis.}
\end{quote}

As with the main tasks, GKMRV is evaluated under both Direct and Chain-of-Thought prompting (Table~\ref{tab:prompt_GKMRV}, and \ref{tab:prompt_cot_GKMRV}), with ten completions per model. By disentangling declarative knowledge from inferential structure and designing tasks that reflect core dimensions of clinical reasoning, this framework enables principled conclusions about LLM performance in CTNLI beyond dataset-specific artifacts.

\subsection{Reasoning Tasks}

\subsubsection{Causal Attribution}

Causal attribution requires evaluating whether a statement expresses a genuine causal claim, as opposed to a purely correlational or observational report. Formally, let $T$ denote a treatment variable (e.g., administration of a drug) and $Y$ an outcome variable (e.g., a clinical measurement such as serum calcium). Following the interventionist framework of Pearl's \emph{do}-calculus~\cite{pearl2009causality}, the causal effect (CE) of $T$ on $Y$ is defined as:
\begin{equation*}
    \text{CE}(T, Y) \triangleq \mathbb{E}[Y \mid \text{do}(T = 1)] - \mathbb{E}[Y \mid \text{do}(T = 0)]
\end{equation*}
Where $\mathbb{E}[\cdot]$ denotes the expectation, or average value, of the outcome variable $Y$ over all individuals or possible world states consistent with the specified intervention. The notation $\text{do}(T = t)$ represents a hypothetical manipulation in which the treatment variable $T$ is set to value $t$ by external intervention, severing all incoming causal dependencies in the underlying structural model. In this scenario, $\mathbb{E}[Y \mid \text{do}(T = 1)]$ is the expected value of $Y$ if all individuals were treated, while $\mathbb{E}[Y \mid \text{do}(T = 0)]$ is the expected value of $Y$ if no individuals were treated. The difference captures the average causal effect of the treatment.

This is distinct from the conditional expectation $\mathbb{E}[Y \mid T = t]$, which captures passive observation and is typically confounded by external variables. For example, observing improved calcium levels in patients who received a drug does not imply that the drug caused the improvement, unless we can account for what would have happened had the drug not been administered.

Causal reasoning therefore requires a structural representational framework composed of composable latent variables, capable of simulating counterfactual outcomes under different treatment assignments. We operationalize this distinction by constructing examples in which the premise includes observational outcomes (e.g., changes in lab values following treatment, or adverse event reports), but the hypothesis makes a causal assertion (e.g., “The treatment was effective”). Agents lacking an explicit and coherent internal representations are expected to conflate statistical association with causal influence, and as a result, may predict entailment when presented with purely observational evidence. The generation template for such examples is shown in Table~\ref{tab:causal-template}.

In the causal attribution setting, the associated GKMRV Task requires the model to assess whether a meta-level statement $\psi'$ correctly identifies why a causal interpretation is invalid for a given premise $\varphi$. Let $\mathcal{K}$ denote general clinical research principles (e.g., the need for control groups). The GKMRV statement is valid iff:
$$
\mathcal{K} \models_{\text{CTNLI}} \psi' \quad \text{given } \varphi
$$
That is, $\psi'$ is entailed by applying domain knowledge $\mathcal{K}$ to the structure of $\varphi$. Unlike the main causal attribution task, which tests inferential validity, this task probes whether the model \emph{knows} that certain study designs preclude causal claims.

This allows us to distinguish factual ignorance from inferential failure. The generation template is shown in Table~\ref{tab:causal-template}.

\begin{table}[h!]
\centering
\small
\begin{tabular}{>{\ttfamily\raggedright\arraybackslash}p{0.9\columnwidth}}
\toprule
\textbf{Premise}: Either \\
(a) Outcome Measurement: [metric] assessed [time-frame] after first dose of [drug]. \\
\quad Results: [p] [Outcome Measurement] \\
(b) Adverse Events 1: Total: [p]/[n] ([x]\%) \\
\quad[event\_1] [k]/[n] ([x]\%) \\
\quad[event\_2] [k]/[n] ([x]\%) \ldots\\ \midrule
\textbf{Statement}: [drug] causes/improves [metric description] \emph{or} The treatment was effective.\\
\textbf{label}: contradiction \\\midrule
\textbf{GKMRV}: (a) Results: [p] [Outcome Measurement] does not prove that [drug] is effective as there is no comparison group \\
(b) Adverse Events do not provide information about how effective a treatment is. Only the presence or absence of side effects.  \\
\textbf{label}: True \\\midrule
\textbf{Control GKMRV}: (a) Results: [p] [Outcome Measurement] proves that [drug] is effective despite the lack of a comparison group \\
(b) Adverse Events provide information about how effective a treatment is. \\
\textbf{label}: False \\
\bottomrule
\end{tabular}
\caption{Template for the causal attribution tasks.}
\label{tab:causal-template}
\end{table}

\subsubsection{Compositional Grounding}

Compositional grounding arises when the truth of a hypothesis $\psi$ cannot be inferred from any single atomic subformula in the premise $\varphi$. Instead, entailment depends on the joint interpretation of multiple predicates \cite{chomsky1972studies} under the interpretation function $I$ in a model $\mathcal{M} = \langle W, \mathfrak{D}, I \rangle$. This occurs when $\psi$ asserts a property or outcome, such as clinical benefit or toxicity, that emerges from structured co-dependence among variables and predicates distributed across $\varphi$.

Co-dependence in first-order logic typically arises through variable binding across multiple predicates, particularly within quantified conjunctions \cite{smith2003introduction}. For example, let \( D(x) \) denote that drug \( x \) is administered, \( Z(x) \) denote that dose \( x \) is administered, and \( T(d, z) \) denote that toxicity occurs under drug \( d \) at dose \( z \). Then the co-dependence of a specific drug \( d_0 \) and dose \( z_0 \) with respect to toxicity is defined as:
\[
T(d_0, z_0) \leftrightarrow \big( D(d_0) \land Z(z_0) \big)
\]
This expresses that toxicity arises if and only if both the specific drug and the specific dose are present. Neither condition alone suffices; the effect is compositionally grounded in their conjunction.

More generally, the truth of a hypothesis asserting a clinical effect must be grounded in a \emph{compositional semantic structure} over a tuple of clinical inputs. Let \( \text{Benefit}(d, z, dx, s) \) denote that a patient with diagnosis \( dx \) experiences clinical benefit when treated with drug \( d \) at dose \( z \) on schedule \( s \). Then the hypothesis is entailed only if the configuration
\[
x := \langle d, z, dx, s \rangle \in \mathfrak{D}_{\text{CT}}^4
\]
is semantically valid under the interpretation function:
\[
I_{\text{CT}}(\text{Benefit})(x) \rightarrow True
\]
In compositional grounding, the interpretation of a predicate such as \( P(x) \) depends not only on its lexical meaning but also on the semantic compatibility of co-occurring predicates in the same premise. Formally, given a premise \( \varphi = P(x) \land Q(y) \land R(z) \), the contribution of \( P(x) \) to the truth of \( \psi \) is evaluated with respect to the composite configuration \( x := \langle x, y, z \rangle \), and whether this configuration satisfies the clinical constraints embedded in the model \( \mathcal{M}_{\text{CT}} = \langle W_{\text{CT}}, \mathfrak{D}_{\text{CT}}, I_{\text{CT}} \rangle \).

For example, let \( P(x) = \text{Administered}(x) \), where \( x \) is a drug. In isolation, \( \llbracket P(\text{Trastuzumab}) \rrbracket^{w} = \text{True} \) might support the hypothesis \( \psi = \text{EffectiveTreatment} \). However, if the full configuration \( \langle \text{Trastuzumab}, 500\text{mg}, \text{HER2-} \rangle \) violates indication or dosing constraints in all worlds \( w \in W_{\text{CT}} \), then \( \llbracket \psi \rrbracket^{w} = \text{False} \). Thus, the semantic contribution of \( P(x) \) is re-contextualized via its interaction with other predicates, and only acquires entailment force if the entire configuration is semantically licensed.

Evaluating this condition requires that the inference model maintain a coherent and explicitly representable model of the clinical domain. This includes:
\begin{itemize}
  \item An internally consistent belief state over clinically admissible worlds \( w \in W_{\text{CT}} \), each reflecting a plausible and protocol-compliant scenario.
  \item Constraint-checking mechanisms in \( I_{\text{CT}} \), including compatibility between diagnosis and treatment, valid therapeutic ranges, scheduling rules, and exclusion criteria.
  \item The capacity to construct and evaluate structured tuples derived from multiple atomic premises, preserving variable bindings and modeling interdependencies.
\end{itemize}

Without such structured, compositional representations, the model cannot determine whether the composite configuration \( x \) is semantically valid under \( I_{\text{CT}} \), nor whether the truth of \( \psi \) follows in any world \( w \in W_{\text{CT}} \). Compositional grounding in CTNLI thus operationally requires both the synthesis of multi-entity configurations and their evaluation against domain-specific semantics and constraints.

We evaluate this capability using the template shown in Table~\ref{tab:compositional-template}, where the model must infer whether the conjunction of treatment, dosage, diagnosis, patient details and scheduling details yields a medically valid configuration that supports the stated clinical outcome. Contradictions arise when surface-level plausibility masks semantic violations detectable only through compositional grounding.

The GKMRV Task for compositional grounding assesses whether the model can recognize that a treatment configuration described in the premise violates established clinical safety or protocol constraints. The associated statement highlights the medical risks or inappropriateness of the regimen, such as excessive cumulative dose, unsafe scheduling, or incompatibility with the patient’s diagnosis.

Rather than inferring a clinical outcome, the model must judge whether the configuration itself is invalid based on standard pharmacologic or oncologic knowledge. Success on this task indicates that the model encodes relevant background knowledge required to flag clinically unacceptable regimens, independent of its ability to reason about downstream effects. The corresponding generation template is shown in Table~\ref{tab:compositional-template}.

\begin{table}[h!]
\centering
\small
\begin{tabular}{>{\ttfamily\raggedright\arraybackslash}p{0.9\columnwidth}}
\toprule
\textbf{Premise}: [age]-year-old [sex] with [diagnosis] receiving [drug] at [dose] [units] [schedule] [optional concurrent drug].\\\midrule
\textbf{Statement}: The patient is expected to experience [clinical benefit] and [effect of drug with standard-of-care dosage].\\
\textbf{label}: contradiction \\\midrule
\textbf{GKMRV}: [drug] at [dose] [units] [schedule] is [contraindicated/harmful] because [toxicity description] \\
\textbf{label}: True \\\midrule
\textbf{Control GKMRV}: [drug] at [dose] [units] [schedule] provides [effect of drug with standard-of-care dosage] with no risk of [toxicity description] \\
\textbf{label}: False \\
\bottomrule
\end{tabular}
\caption{Template for the compositional grounding task.}
\label{tab:compositional-template}
\end{table}

\subsection{Epistemic Verification}

Epistemic Verification is a distinctive challenge in CNLI, all input propositions in the premise are epistemically mediated: they reflect recorded or reported assertions, not direct truths accessible to the inference model. Unlike some general-domain NLI tasks where statements are interpreted as if from a first-person omniscient narrator (with access to ground truth), CTNLI must treat all statements as second-hand, potentially fallible, and anchored to a specific speaker or source \cite{van2011logical, fagin2004reasoning}. Formally, this aligns with the notion that propositions in $\mathcal{L}_{\text{CT}}$ carry implicit \emph{factivity predicates} such that $\varphi$ denotes not simply a proposition $p$, but a commitment $K_a(p)$ by an agent $a$ (e.g., a clinician, patient, or protocol record) \cite{fagin2004reasoning}. As such, assertions in CTNLI must be treated as \emph{defeasible}: they are presumed true unless contradicted by stronger evidence (e.g., imaging data, biomarker measurements) or rendered implausible by clinical constraints (e.g., timing inconsistencies or ontology violations). 

We therefore model each premise $\varphi \in \mathcal{L}_{\text{CT}}$ as a conjunction of individual epistemic commitments $K_a(\varphi_i)$ by some source agent $a$:
\[
\varphi = K_a(\varphi_1) \land K_a(\varphi_2) \land \dots \land K_a(\varphi_n)
\]
A premise exhibits \emph{internal inconsistency} if it contains two or more embedded assertions that cannot jointly be true in any clinically admissible world. Formally, $\varphi$ is inconsistent if $\exists \varphi_i, \varphi_j \subseteq \varphi,\;\forall w \in W_{\text{CT}}$ such that:
\[\;
\llbracket \varphi_i \rrbracket^{\mathcal{M}_{\text{CT}}, g, w}\rightarrow True
\Rightarrow 
\llbracket \varphi_j \rrbracket^{\mathcal{M}_{\text{CT}}, g, w}\rightarrow False
\]
In this case, the set of assertions $\{\varphi_1, \dots, \varphi_n\}$ does not describe a coherent or logically satisfiable clinical scenario. The model $\mathcal{M}_{\text{CT}}$ must therefore treat the premise as epistemically unreliable, and at least one $K_a(\varphi_i)$ is incorrect. For example, consider the statements ``Weight: 150kg'' and ``BMI: 17'' in a premise. Under clinical constraints encoded in $I_{\text{CT}}$, this pair cannot jointly hold for any plausible height value i.e., no $w \in W_{\text{CT}}$ satisfies both $\llbracket \text{Weight} = 150 \rrbracket^w$ and $\llbracket \text{BMI} = 17 \rrbracket^w$. In these cases the model must determine which individual commitments $K_a(\varphi_i)$ to preserve and which to reject. To do so, we introduce a plausibility or credibility function:
\[\pi : \mathcal{L}_{\text{CT}} \rightarrow \mathbb{R}\]
where $\pi(\varphi_i)$ encodes the relative epistemic reliability of each assertion $\varphi_i$, as determined by source attribution, semantic content, and clinical domain constraints. Assertions with lower plausibility are more likely to be revised or rejected in the face of inconsistency. Suppose $\varphi_i, \varphi_j \subseteq \varphi$ such that:
\[
\forall w \in W_{\text{CT}},\; 
\llbracket \varphi_i \rrbracket^{w} = True \Rightarrow \llbracket \varphi_j \rrbracket^{w} = False,
\]
Where $\pi(\varphi_i) < \pi(\varphi_j)$. Then $K_a(\varphi_i)$ is rejected to restore consistency. In some cases, an assertion $\varphi_i$ may be implausible in isolation, i.e.,
\[
\forall w \in W_{\text{CT}},\; \llbracket \varphi_i \rrbracket^{w} = False
\]
For example ``Weight: 1200kg'' for a human subject. Here, $\pi(\varphi_i)$ is near zero by virtue of violating ontological or physiological constraints embedded in $I_{\text{CT}}$, and the assertion is rejected outright as clinically invalid. When $\pi(\varphi_i) \approx \pi(\varphi_j)$ and no clear preference can be established, the model must abstain from inference (i.e., label the hypothesis as \textsc{neutral}).

We operationalise this epistemic verification challenge by constructing examples in which the premise contains an asserted diagnosis, while the hypothesis asserts an alternative diagnosis better supported by the evidence. Agents lacking coherent and composable internal representations capable of evaluating internal inconsistency, assertion plausibility and resolving conflicting commitments are expected to conflate speaker commitment with ground truth. As a result, they may incorrectly predict entailment for clinically implausible or unsupported assertions. The generation template for such examples is shown in Table~\ref{tab:epistemic}.

The epistemic GKMRV Task evaluates whether the model can identify that a reported clinical assertion lacks plausibility or evidential support, independently of resolving an entailment relation. The premise presents a clinician-asserted diagnosis or interpretation, while the associated ground knowledge statement challenges its validity based on inconsistencies, insufficient evidence, or ontological implausibility.

Rather than choosing between competing conclusions, the model must determine whether the reported claim violates clinical reasoning norms e.g., diagnosing sepsis without infection markers, or claiming remission despite progressive imaging. This task isolates whether the model encodes the necessary background knowledge to recognize implausible or unsupported medical conclusions, even when those conclusions are asserted by a presumed expert. The corresponding generation template is shown in Table~\ref{tab:epistemic}.

\begin{table}[h!]
\centering
\small
\begin{tabular}{>{\ttfamily\raggedright\arraybackslash}p{0.9\columnwidth}}
\toprule
\textbf{Premise}: [clinical findings]. A [role] diagnoses [incorrect diagnosis] based on [inadequate evidence].\\\midrule
\textbf{Statement}: Either \\
(a) The patient has [incorrect diagnosis]. \\
(b) The patient has [correct diagnosis]. \\
\textbf{label}: (a) contradiction; (b) entailment \\\midrule
\textbf{GKMRV}: [incorrect diagnosis] is [not supported/contradicted] by [clinical findings] and [inadequate evidence] \\
\textbf{label}: True \\\midrule
\textbf{GKMRV}: [incorrect diagnosis] is [supported] by [clinical findings] despite [inadequate evidence] \\
\textbf{label}: False \\
\bottomrule
\end{tabular}
\caption{Template for the Epistemic Verification task.}
\label{tab:epistemic}
\end{table}

\subsubsection{Risk State Abstraction}

Risk is the exposure to the possibility of loss, injury, or other adverse or unwelcome circumstance. In a formal semantic framework, where the truth of clinical claims is evaluated with respect to possible worlds \( w \in W_{\text{CT}} \), we define the risk state abstraction of \(Risk(\varphi, \psi)\) as:
\[
\mathbb{E}_{w \sim \Pr(w \mid \varphi)} 
\left[ 
\sum_{e \in \mathcal{E}(w, \psi)} 
\Pr(e \mid w) \cdot \mathcal{A}(e, w) 
\right]
\]
This formulation is inspired by expected utility models in medical decision theory \cite{hunink2014decision}, where risk is quantified as the product of an event’s likelihood and its adverse consequence. Where \( \Pr(w \mid \varphi) \) is the posterior distribution over clinically admissible worlds \( w \in W_{\text{CT}} \), conditioned on the premise \( \varphi \). Constructing this distribution requires a coherent and explicit internal model of the clinical domain. The model must infer which events, conditions, or trajectories are consistent with the available information and not ruled out by \( \varphi \), while also assigning appropriate probabilities based on incomplete or indirect evidence.
\vspace{0.5em}
\noindent
\begin{itemize}
    \item \( \mathcal{E}(w, \psi) \) is the set of clinically relevant events in world \( w \) that bear on the truth of the hypothesis \( \psi \). These events may not be explicitly stated in the premise or hypothesis, and identifying them demands a model capable of recalling and reasoning over plausible latent outcomes e.g., unconfirmed diagnoses, adverse reactions, or deteriorating conditions.
    \item \( \Pr(e \mid w) \) is the probability of event \( e \) occurring in world \( w \). Accurate estimation requires the model to track conditional dependencies within \( w \), such as the likelihood of disease progression or treatment failure, conditioned on clinical findings.
    \item \( \mathcal{A}(e, w) \in \mathbb{R}_{\geq 0} \) is the adverse outcome function, representing the contextual magnitude of harm resulting from event \( e \) in world \( w \). Estimating this value often necessitates simulating counterfactual causal chains i.e., what would happen if the condition were left untreated or mismanaged—making it sensitive to both temporal and interventional dynamics within \( w \).
\end{itemize}

We evaluate a model's ability to perform risk abstraction using two templates (Table~\ref{tab:risk}). In setting (a), the premise explicitly presents adverse event frequencies, and the task is to identify which event poses greater clinical risk. This tests whether the model correctly integrates event probability and adverse outcome magnitude, particularly when these are in tension.

Setting (b) probes latent risk reasoning: the premise presents findings suggestive of serious but unconfirmed conditions, while the hypothesis asserts a management action (e.g., discharge). The model must identify un-ruled-out adverse events, estimate their posterior likelihood, and assess the harm if left untreated. This requires a coherent internal architecture of composable latent representations of concepts, capable of simulating causal trajectories under uncertainty.

In both cases, the model must evaluate the hypothesis \( \psi \) not as a purely logical consequence of \( \varphi \), but as a summary of the expected clinical risk it entails.

In the risk state abstraction setting, the GKMRV Task evaluates whether the model can recognize and articulate why a given clinical scenario entails elevated or unacceptable patient risk. The associated statement $\psi'$ identifies the risk factor that warrants intervention, escalation, or alternative clinical management. The GKMRV probes whether the model encodes the underlying principles of risk stratification—such as the severity of thrombosis relative to common but benign side effects, or the danger of outpatient management in neutropenic fever. This task isolates whether the model has access to latent knowledge about risk magnitude, thresholds for intervention, and plausible harm trajectories. The corresponding generation template is shown in Table~\ref{tab:risk}.

\begin{table}[h!]
\centering
\small
\begin{tabular}{>{\ttfamily\raggedright\arraybackslash}p{0.9\columnwidth}}
\toprule
\textbf{Premise}: Either \\
(a) Adverse Events: Total: [p]/[n] ([x]\%) \\
\quad [minor event] [k\textsubscript{minor}]/[n] ([x]\%) \\
\quad [severe event] [k\textsubscript{severe}]/[n] ([x]\%) \quad \ldots \emph{([k\textsubscript{severe} < k\textsubscript{minor}])} \\
(b) EHR: [clinical findings suggestive of severe underlying condition] [extraneous contextual details] \\\midrule
\textbf{Statement}: Either \\
(a) The highest patient risk was associated with [minor event] \\
(b) [clinical management decision] \\
\textbf{label}: contradiction for (a); entailment for (b) \emph{iff} decision reflects optimal risk reasoning; otherwise contradiction. \\\midrule
\textbf{GKMRV}: \\ (a) Despite [k\textsubscript{severe} < k\textsubscript{minor}] [severe event] represents a higher patient risk than [minor event] due to its greater potential for serious harm. \\
(b) [severe underlying condition] is not ruled out by [clinical findings suggestive of severe underlying condition] and represents significant risk of serious harm if left untreated.  \\
\textbf{label}: True \\\midrule
\textbf{GKMRV}: \\ (a) As [k\textsubscript{severe} < k\textsubscript{minor}] [minor event] represents a higher patient risk than [severer event] regardless of its greater potential for serious harm. \\
(b) [severe underlying condition] is completely ruled out by [clinical findings suggestive of severe underlying condition] and represents no significant risk of serious harm if left untreated.  \\
\textbf{label}: False \\
\bottomrule
\end{tabular}
\caption{Template for the Risk State Abstraction task.}
\label{tab:risk}
\end{table}

\begin{table}[t]
\centering
\scriptsize
\begin{tabular}{l@{\hskip 4pt}c@{\hskip 4pt}c@{\hskip 4pt}c@{\hskip 4pt}c}
\toprule
\textbf{Model} & \textbf{Causal Attr.} & \textbf{Comp. G.} & \textbf{Epis. Verif.} & \textbf{Risk Abstr.} \\
\midrule
\multicolumn{5}{l}{\textit{CoT}} \\[-0.4em]
\midrule
deepseek-r1     & 0.50 (1.00) & \textbf{0.19} (1.00) & \textbf{0.31} (1.00) & 0.13 (0.87) \\
gemini 2.5      & \textbf{0.73} (0.99) & 0.01 (1.00) & 0.29 (1.00) & 0.11 (0.94) \\
o3              & 0.72 (1.00) & 0.00 (0.97) & 0.17 (0.98) & 0.00 (0.86) \\
gpt4o           & 0.24 (1.00) & 0.02 (1.00) & 0.29 (1.00) & 0.05 (0.94) \\
gpt4o-mini      & 0.64 (0.95) & 0.00 (0.96) & 0.08 (0.95) & 0.01 (0.81) \\
llama3.2        & 0.29 (0.68) & 0.10 (0.79) & 0.27 (0.79) & \textbf{0.27} (0.66) \\
\midrule
\multicolumn{5}{l}{\textit{Direct}} \\[-0.4em]
\midrule
deepseek-r1     & 0.31 (1.00) & \textbf{0.14} (0.99) & 0.32 (1.00) & 0.16 (0.88) \\
gemini 2.5      & \textbf{0.77} (0.97) & 0.02 (0.99) & 0.28 (0.99) & 0.07 (0.92) \\
o3              & 0.73 (1.00) & 0.00 (1.00) & 0.18 (1.00) & 0.00 (0.90) \\
gpt4o           & 0.32 (1.00) & 0.02 (1.00) & 0.23 (0.99) & 0.03 (0.89) \\
gpt4o-mini      & 0.22 (1.00) & 0.00 (0.96) & 0.15 (0.88) & 0.05 (0.82) \\
llama3.2        & 0.07 (0.56) & 0.05 (0.87) & \textbf{0.33} (0.58) & \textbf{0.25} (0.81) \\
\bottomrule
\end{tabular}
\caption{
Accuracy on each reasoning task with corresponding GKMRV accuracy shown in parentheses.}
\label{tab:prompting_rows}
\end{table}

\begin{figure*}[t]
    \centering
    \includegraphics[width=0.75\textwidth]{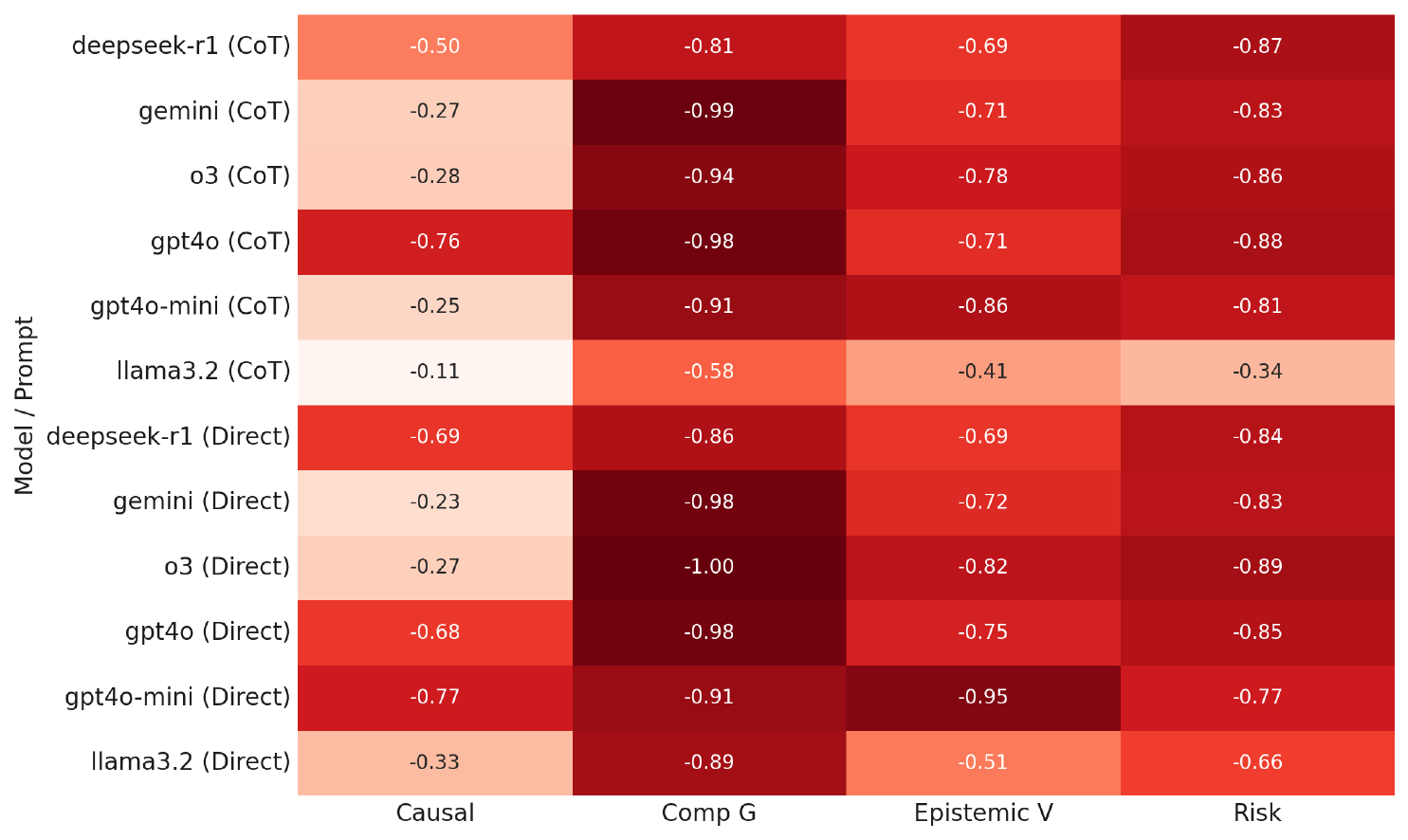} 
    \caption{
        Accuracy gap between main task performance and GKMRV accuracy, across four tasks. Values closer to zero indicate strong alignment between reasoning and underlying knowledge.}
    \label{fig:accuracy-gap-heatmap}
\end{figure*}

\begin{table}[h]
\centering
\scriptsize
\begin{tabular}{p{4cm}cc}
\toprule
\textbf{Reasoning Task} & \textbf{Consistency} \\
\midrule
Causal Attribution & 0.84 \\
Compositional Grounding & 0.87\\
Epistemic Verification & 0.90 \\
Risk Abstraction & 0.85 \\
\bottomrule
\end{tabular}
\caption{Consistency, defined as the ratio of majority label count to total responses across completions per example, averaged across prompt types and models. Higher values (closer to 1) indicate greater agreement among completions.}
\label{tab:consistency}
\end{table}

\section{Overview of Empirical Evidence of Fundamental Limitations}
Across all four reasoning tasks; Causal Attribution, Compositional Grounding, Epistemic Verification, and Risk State Abstraction, \textbf{our results reveal a systematic and domain-general dissociation between declarative knowledge retrieval and structured inferential reasoning in current language models} (Figure~\ref{fig:accuracy-gap-heatmap}). 

While models consistently demonstrate near-ceiling performance on the GKMRV tasks with a mean accuracy of 0.918, their main task accuracy remains strikingly low (mean: 0.25), with some tasks exhibiting near-total failure (e.g., Compositional Grounding, avg. accuracy: 0.04), as shown in Table~\ref{tab:prompting_rows}. This discrepancy is not attributable to stochasticity or label noise: model responses exhibit high internal consistency (mean: 0.87), indicating that errors are systematic and reproducible. 

In particular, models appear to possess the relevant clinical facts, diagnostic criteria, and pharmacologic constraints, yet repeatedly fail to integrate these into structured and correct reasoning. Instead, completions reflect the deployment of rigid, often domain-inappropriate heuristics, such as equating numerical effect size with efficacy, decomposing joint constraints into pairwise associations, or deferring to source assertions over evidentiary coherence. These findings suggest a structural limitation in the architecture and inference procedures of contemporary models: the absence of structured, concept-level latent representations that support simulating counterfactuals, resolving internal inconsistencies, and reasoning over compositional or probabilistic configurations. 

The gap between factual retrieval and inferential deployment is both pervasive and quantifiable, and reflects a fundamental barrier to the reliable use of LLMs in clinical reasoning tasks requiring grounded, context-sensitive judgment.

\section{Causal Attribution Analysis}

\paragraph{Near-ceiling GKMRV ($\geq$ 0.95) fails to yield reliable Causal Attribution accuracy (avg. 0.49)} Table~\ref{tab:prompting_rows} reports aggregate performance on the Causal Attribution task, with maximum main task accuracy observed for Gemini under direct prompting (0.77), an average accuracy of 0.49 across all model–prompt pairs, and a minimum of 0.07 for LLaMA 3.2 (direct). In contrast, GKMRV accuracy is near-ceiling: all models achieve above 0.95, with the sole exception of LLaMA 3.2 (0.68). This sharp dissociation suggests that the models possess the necessary background knowledge to assess causal validity, yet fail to operationalize it when evaluating entailment. The failure mode, then, is not epistemic but inferential. Notably, Causal Attribution yields the highest overall and average accuracy of any task in the benchmark, both for the main reasoning task and for GKMRV. This suggests that causal reasoning is better supported than other forms of clinical inference, perhaps due to stronger alignment between causal phrases and lexical patterns learned during pretraining.

\paragraph{Identical single-arm evidence yields divergent causal conclusions, driving lowest task consistency (84\%)} A prototypical case is shown below. The premise describes a single-arm outcome, where no comparator is available:

\begin{quote}\small
\textbf{Premise:} \emph{Outcome: Rate of wound healing at Day 14 following topical application of silver sulfadiazine. 40\% (8/20) of wounds showed full epithelialization by Day 14.}

\textbf{Hypothesis:} \emph{Silver sulfadiazine accelerates wound healing within two weeks.}
\end{quote}

Despite the lack of a control group, and thus no counterfactual comparison, model completions diverge. One chain-of-thought response correctly abstains from causal inference:
\begin{quote}\small
\emph{``The premise shows healing occurred by Day 14 but does not provide a baseline or control group. Thus, acceleration cannot be inferred.''} \texttt{output: neutral}
\end{quote}

However, another completion for the same input asserts a causal conclusion:
\begin{quote}\small
\emph{``40\% of wounds healed by Day 14, which supports the claim of accelerated healing.''} \texttt{output: entailment}
\end{quote}

This divergence is not an isolated artifact. On average, 84\% of model completions on the Causal Attribution task agreed on the same label for a given input (Table~\ref{tab:consistency}), the lowest consistency across all tasks. Importantly, when the same causal principle is evaluated outside the entailment setting, in GKMRV, framed as a direct True/False judgment about the validity of a causal claim, model responses are almost uniformly correct:

\begin{quote}\small
\textbf{Statement:} \emph{This outcome does not prove that silver sulfadiazine is effective because there is no comparison group.} \texttt{output: True}\\
\textbf{Reasoning:} ``The premise describes a single-arm study without a control group. Without a comparator, it is not possible to isolate the effect of the treatment from natural healing, placebo effects, or \ldots'' \texttt{output: True}
\end{quote}

This contrast is revealing. The relevant declarative knowledge is accessible, models can recognize the epistemic limitation when it is explicitly foregrounded. Yet this knowledge fails to translate into consistent entailment behavior.

\paragraph{Causal heuristic drift: numeric effects misread as proof of efficacy, replacing interventional reasoning} Across completions, we observe an alternation between sound causal logic and shallow lexical heuristics. In particular, models often equate the presence of a numerical effect (e.g., a proportion of responders) with proof of efficacy. Rather than reasoning over counterfactuals or comparing potential outcomes, they default to a pattern-matching heuristic: \(\llbracket \varphi \rrbracket = \) \(\text{``}\alpha\% "responded'' \) \(\;\rightarrow\;\) \(\llbracket \psi \rrbracket = ``effective''\). Thereby collapsing the epistemic modal $\operatorname{do}$ to an observational conditional. In the majority of responses, there is no attempt to estimate or even approximate the true causal effect. Even when the model produces the correct label (e.g., \textsc{neutral}), the reasoning is often degenerate, rejecting entailment not due to the absence of a counterfactual comparison, but because the observed response rate (e.g., 40\%) is deemed too low to be clinically meaningful. This suggests that the model’s decision boundary is not anchored in interventional semantics, but in loose thresholds over lexicalized effect sizes.

These results point to what we term \emph{causal heuristic drift}: model behaviour reflects shifts in local salience e.g., over “40\% healed” and “accelerates”, rather than stable, structure-sensitive reasoning. The resulting outputs are unmediated by formal semantics, and even plausible causal inferences are revealed to be byproducts of emergent pattern completion, not evidence of structured deliberation.

Critically, the models do appear to encode the causal principle in question (e.g., the need for a comparator), but fail to apply it reliably during inference. In effect, they know the rule, but do not know when or how to use it. This dissociation between declarative knowledge and inferential deployment highlights a structural deficit: reliable causal inference appears to require systematic reasoning grounded in a coherent internal representation of interventions and outcomes. The observed failures indicate that current models lack such a representation, and therefore cannot support robust causal judgments, even when the requisite knowledge is nominally available. Plausible outputs, in this light, reflect surface fluency rather than principled generalization.

\section{Compositional Grounding Analysis}

\paragraph{Near-ceiling GKMRV on drug–dose–condition validity ($\geq$0.96) fails to yield Compositional Grounding accuracy (avg. 0.04)}
Table~\ref{tab:prompting_rows} reports aggregate performance on the Compositional Grounding task. Maximum main task accuracy is 0.19 (DeepSeek-R1, CoT), while the lowest is 0.00, observed for multiple models (GPT-4o-mini, o3, GPT-4o). The average accuracy across all model–prompt pairs is just 0.04, indicating a near-total failure to perform the task despite a consistency rate of 87\% (Table~\ref{tab:consistency}), suggesting that models are confidently and repeatedly wrong in a structurally systematic manner. In contrast, GKMRV performance is uniformly high, with all model–prompt pairs achieving $\geq$0.96 with the exception of LLaMa 3.2. This discrepancy indicates that models possess factual knowledge about drug–dose–condition compatibility, yet fail to integrate these facts compositionally at inference time. This discrepancy further reinforces that the failure is not epistemic but structural.

\paragraph{Identical high-risk prescription prompts opposing judgments, despite clear contraindication}
A representative case involves a high-dose prescription of metformin to a patient with severe renal impairment:
\begin{quote}\small
\textbf{Premise:} \emph{68-year-old female with type 2 diabetes and chronic kidney disease (eGFR 25 mL/min/1.73m$^2$) receiving metformin 1000mg twice daily.}

\textbf{Hypothesis:} \emph{The treatment is expected to improve glycemic control, reduce cardiovascular risk, and provide long-term kidney protection.}
\end{quote}

One response correctly identifies the contraindication:
\begin{quote}\small
\emph{``Metformin is contraindicated in patients with severe renal impairment due to increased risk of lactic acidosis, especially at high doses like 1000mg twice daily...''} \texttt{output: contradiction}
\end{quote}

Yet another response to the same input fails to register the clinical risk:
\begin{quote}\small
\emph{``Metformin is commonly used for glycemic control... but it is not known for providing long-term kidney protection nor specifically reducing cardiovascular risk on its own... the relationship appears to be neutral.''} \texttt{output: neutral}
\end{quote}

In contrast, the corresponding GKMRV task, framed as a direct truth judgment, elicits consistent and well-formed responses:

\begin{quote}\small
\textbf{Statement:} \emph{This dosage of metformin is contraindicated because severe renal impairment (eGFR <30) greatly increases the risk of lactic acidosis}

\textbf{Reasoning:} \emph{``According to established medical guidelines an eGFR below 30 mL/min/1.73m$^2$ is classified as severe renal impairment... significantly increases the risk of a rare but potentially fatal side effect known as metformin-associated lactic acidosis...''} \texttt{output: True}
\end{quote}

\paragraph{Models fail to represent joint clinical constraints, defaulting to decomposed pairwise associations} The discrepancy highlights a failure to represent and reason over \emph{joint clinical constraints}. Instead of evaluating the ternary relation $\texttt{EffectiveAtDose}(x, y, z)$, capturing drug $x$ at dose $y$ for condition $z$, models default to decomposed associations like $\langle x, z \rangle \in I(\texttt{Effective})$ and $\langle x, y \rangle \in I(\texttt{ReasonableDose})$. Failing to recognise these interdependencies leads to endorsements of drug–dose pairs that are formally contraindicated. The inference engine thereby substitutes structured semantic evaluation with independent lexical associations. 

\paragraph{Surface-level lexical matches drive 87 \% consistent yet systematically wrong outputs}
The observed 87\% consistency in model predictions (Table \ref{tab:consistency}, despite an average task accuracy of only 0.04, strongly suggests that the few correct responses are products of occasional alignment between surface-level token associations and the correct label—an artifact of surface fluency rather than structured inference.

Despite possessing declarative knowledge of clinical contraindications, models fail to engage in genuine grounded compositional reasoning. The deficit is not attributable to an absence of clinical facts in the training data; rather, the models fail to recognise those facts must be compositionally integrated to support correct reasoning. As with causal inference, the failure is not due to gaps in ground knowledge but to the absence of structured reasoning mechanisms capable of composing clinical facts into coherent judgments. 

Thus, while models clearly encode declarative knowledge of clinical contraindications, they fail to integrate that knowledge in contextually appropriate ways. The deficit is not epistemic but structural: correct reasoning requires the composition of clinical facts into coherent, context-sensitive judgments—a capacity current models demonstrably lack.

\section{Epistemic Verification Analysis}
\paragraph{Near-ceiling GKMRV on diagnostic validity ($\geq$0.95) contrasts with low Epistemic Verification accuracy (avg. 0.24)}
Table~\ref{tab:prompting_rows} reports model performance on the Epistemic Verification task. Accuracy remains uniformly low across models and prompting strategies, with a maximum of 0.33 achieved by LLaMA 3.2 under direct prompting and a cross-model mean of 0.24. As with prior tasks, these low accuracy scores are accompanied by high internal consistency: 90\% of completions assign the same label to a given input across runs and prompts (Table~\ref{tab:consistency}). In line with the broader pattern observed across reasoning tasks, the models appear to apply rigid heuristics in place of context-sensitive inference, yielding highly reproducible yet frequently incorrect outputs.

\paragraph{The same clinical evidence yields contradictory labels due to unstable evidential reasoning}
A representative failure is shown below. The premise presents clinically specific evidence:

\begin{quote}\small
\textbf{Premise:} \emph{A 25-year-old female presents with fever, flank pain, and dysuria. Urine culture is positive for \textit{E. coli}. Despite this, the physician suggests that the pain is likely due to a pulled muscle and no antibiotics are prescribed.}

\textbf{Hypothesis:} \emph{The patient has a UTI.}
\end{quote}

Under standard diagnostic criteria, the presence of dysuria, flank pain, fever, and a positive urine culture for E. coli constitutes a strong basis for diagnosing a urinary tract infection (UTI), likely pyelonephritis. The hypothesis is therefore \textsc{entailed}, as it follows from the premise in all clinically admissible worlds $w \in W_{\text{CT}}$. Under the interpretation function $I_{\text{CT}}$, these findings have high diagnostic specificity. This interpretation is corroborated by specific model completions that correctly prioritize evidential coherence over deference to authority.
\begin{quote}\small
\emph{``The urine culture result showing positive for \textit{E. coli} is a classic indication of a UTI... Despite the physician’s suggestion, the symptoms and test results strongly support the diagnosis.”} \texttt{output: entailment}
\end{quote}

This output suggests sensitivity to the evidentiary structure underlying diagnostic inference. However, this reasoning is not applied consistently. Again, alternate completions for the same input yield mutually inconsistent interpretations:
\begin{quote}\small
\emph{``The physician suggests an alternative cause... Since the statement seems reasonable based on objective evidence but is contradicted by the physician's assessment, the relationship is a contradiction.”} \texttt{output: contradiction}
\end{quote}

\paragraph{Clinician-overrule heuristics override strong objective evidence, breaking epistemic coherence}
The phrase \textit{“reasonable based on objective evidence but}” exemplifies a fundamental breakdown in epistemic reasoning, where contradictory cues are treated as jointly plausible. Rather than adjudicating among conflicting propositions based on evidential strength, the model defaults to a heuristic that treats the clinician’s suggestion as an epistemic override. Formally, while the model is capable of identifying the \emph{internal inconsistency} within the premise, it frequently fails to resolve the inconsistency through evidential reasoning. Additionally, we observe a tendency to treat clinician assertions \( K_{\text{clinician}}(\varphi_i) \) as having maximal plausibility, even when directly contradicted by strong observational evidence \(\varphi' \models \lnot \varphi \). This epistemic deference is not applied reliably, but when invoked, it systematically subverts clinical logic.

Main task performance remains low, despite GKMRV accuracy again being at ceiling for all models except LLaMA 3.2. The models clearly possess the relevant declarative knowledge (e.g., symptom patterns, diagnostic criteria), as demonstrated by the following high-fidelity response to the corresponding GKMRV statement:

\begin{quote}\small
\textbf{Statement:} \emph{The suggestion of a pulled muscle is incorrect because the positive urine culture with typical symptoms confirms a urinary tract infection.} \texttt{output: True}
\end{quote}

These results parallel earlier findings, despite high internal consistency (90\%), and near perfect performance on the GKMRV task, without a structured mechanism to weigh and resolve competing propositions, models default to heuristics that mimic fluency with a fundamental lack of understanding of evidential coherence and diagnostic plausibility.

\section{Risk State Abstraction Analysis}

\paragraph{GKMRV performance on risk validity high (avg. 0.86) but fails to transfer to Risk State Abstraction ($<$ 0.27 accuracy)}
Model performance on Risk State Abstraction is uniformly poor, with all model–prompt pairs achieving less than 0.27 accuracy (Table~\ref{tab:prompting_rows}). Despite the low accuracy, internal consistency remains relatively high at 0.85 (Table~\ref{tab:consistency}). These findings parallel the trends observed in prior tasks. 

\paragraph{Failure to activate latent high-risk diagnoses blocks correct urgent-evaluation judgments}
A representative failure is shown below:
\begin{quote}\small
\textbf{Premise.} \emph{Patient (52M), currently on holiday, presents with several months' history of nasal congestion (blocked, crusty), generalized fatigue, and widespread myalgia. Reports feeling generally unwell.}

\textbf{Hypothesis.} \emph{The patient requires urgent outpatient evaluation.}
\end{quote}

The correct label is \textsc{entailment}, as this presentation is suggestive of Granulomatosis with Polyangiitis (GPA), a rare vasculitis associated with substantial morbidity and mortality. From a clinical reasoning perspective, latent event inference $\mathcal{E}(w, \psi)$ should activate GPA as a candidate explanation in plausible worlds $w \sim \Pr(w \mid \varphi)$, and the downstream utility $\mathcal{A}(e, w)$ of early specialist evaluation is high. However, models were unable to instantiate the set of clinically relevant events $\mathcal{E}(w, \psi)$ implicated by the hypothesis. As a result, it cannot proceed to evaluate either the posterior probability $\Pr(e \mid w)$ or the adverse outcome function $\mathcal{A}(e, w) \in \mathbb{R}_{\geq 0}$. The entire risk abstraction pipeline fails at the initial step, with no activation of plausible latent events that could mediate entailment. 

\paragraph{“Unlikely” conflated with “ruled out,” ignoring nonzero probability and potential harm}
This same structural failure is evident in cases where the adverse event $e$ is \emph{explicitly stated} in $\psi$. Consider:

\begin{quote}\small
\textbf{Premise:} \emph{A 40-year-old female presents with mild fever and pain localized to the renal angle. She denies coughing and has a normal pulse. She suspects pyelonephritis rather than a pulmonary condition.}

\textbf{Hypothesis:} \emph{Pneumonia is ruled out for this patient.}
\end{quote}

The correct label is \textsc{contradiction}. Although pneumonia is less likely given the symptom profile, it cannot be definitively ruled out without imaging or further examination. Formally, the probability of pneumonia $\Pr(e \mid w)$ remains nonzero under $w \sim \Pr(w \mid \varphi)$, and given the high $\mathcal{A}(e, w)$ associated with missed diagnosis, its exclusion from the differential is inappropriate. Yet model completions simplify this reasoning:

\begin{quote}\small
\emph{``These are not indicative of pneumonia... pneumonia indeed seems unlikely... Thus, the statement logically follows.''} \texttt{output: entailment}
\end{quote}

This output demonstrates a collapse of probabilistic abstraction. The model conflates “unlikely” with “ruled out,” and fails to reason over potential harms. However, in the corresponding GKMRV task, the model exhibits perfect performance:

\begin{quote}\small
\textbf{Statement:} \emph{Pneumonia cannot be completely ruled out; although less likely, lower lobe involvement remains a potential risk given the location of the pain and should warrant evaluation due to the potential for serious harm if left untreated.} \texttt{output: True}
\end{quote}

This discrepancy again underscores that the model possesses the declarative facts necessary for correct inference, but lacks a mechanism for structured integration. It does not estimate $\Pr(e \mid w)$ nor $\mathcal{A}(e, w)$, instead substituting heuristics grounded in lexical proximity.

\paragraph{Risk assessment reduced to event frequency, ignoring severity weighting}
The same deficit arises in trade-offs between frequency and severity:
\begin{quote}\small
\textbf{Premise:} \emph{Adverse Events 7: Total: 50/70 (71.43\%)\
Injection site pain 20/70 (28.57\%)\
Myocardial infarction 3/70 (4.29\%)\
Hypertensive crisis 5/70 (7.14\%)\
Seizure 2/70 (2.86\%)\
Dizziness 3/70 (4.29\%)\
Nausea 2/70 (2.86\%).}

\textbf{Hypothesis:} \emph{The highest patient risk was associated with the occurrence of injection site pain.}
\end{quote}

The correct label is \textsc{contradiction}: injection site pain is frequent but minor, whereas myocardial infarction, though rarer, has far greater severity. Risk must be evaluated via the product of $\Pr(e \mid w)$ and $\mathcal{A}(e, w)$. One chain-of-thought response accurately captures this:

\begin{quote}\small
\emph{“Higher risk typically refers to more severe consequences or a higher likelihood of danger... Events like myocardial infarction or hypertensive crisis generally imply higher patient risk... the statement cannot be concluded as true.”} \texttt{output: contradiction}
\end{quote}

But another yields:
\begin{quote}\small
\emph{“Since the percentage for injection site pain is the highest... the statement is supported by the data.”} \texttt{output: entailment}
\end{quote}

This completion substitutes raw frequency for clinical risk, treating $\mathcal{A}(e, w)$ as uniform. It demonstrates that the model lacks an internal abstraction for severity or harm. In summary, the models fail across both axes of risk state abstraction: they neither simulate future harms from latent diagnoses nor integrate frequency with severity to assess event risk.

Despite these failures on the main task, model performance on the associated GKMRV task is markedly higher, with an average accuracy of 0.86. This confirms again that models encode the relevant medical principles, but fail to operationalize them compositionally at inference time. For instance, the following two statements were correctly classified by nearly all models:

\begin{quote}\small
\textbf{Statement:} \emph{Pneumonia cannot be completely ruled out; although less likely, lower lobe involvement remains a potential risk...} \texttt{output: True}

\emph{“The renal angle pain could reflect referred pain from lower-lobe pneumonia. While less likely, pneumonia cannot be excluded. Serious conditions must be ruled out.”}
\end{quote}

\begin{quote}\small
\textbf{Statement:} \emph{Although more patients experienced injection site pain, myocardial infarction carries a far higher risk...} \texttt{output: True}

\emph{“Myocardial infarction is life-threatening. Injection site pain is mild. The statement correctly contrasts frequency with severity.”}
\end{quote}

\paragraph{Correct GKMRV reasoning absent in compositional integration, leaving shallow salience-driven outputs} As with other reasoning tasks, performance on GKMRV probes is decoupled from task accuracy, indicating that declarative knowledge alone is insufficient. The models lack composable latent representations that would support causal simulation, severity estimation, and reasoning over probabilistic outcomes. Instead, they rely on shallow heuristics driven by statistical salience. Consequently, correct predictions—when they occur, are best interpreted as surface-level alignments rather than evidence of principled, domain-grounded reasoning.

\section{Directions for Future Work}

Our empirical findings underscore a fundamental representational limitation of current large language models: despite robust encoding of clinical knowledge, they fail consistently in structured inferential reasoning tasks due to the lack of coherent, composable latent representations. Addressing this deficiency demands targeted research initiatives across multiple complementary directions:

\paragraph{Neuro-symbolic Integration.}
A promising avenue involves the explicit integration of symbolic reasoning frameworks with neural models, known as neuro-symbolic approaches \cite{nawaz2025review}. These frameworks combine the representational flexibility and data-driven optimization of neural networks with the structured, compositional reasoning capacity of symbolic models \cite{calanzone2024logically}. Future work should explore architectures such as neuro-symbolic program synthesis \cite{zhang2025formalizing}, logic-enhanced neural inference engines \cite{quan2024enhancing}, or differentiable symbolic reasoning layers embedded within transformer architectures \cite{tran2025reasoning}. Such integrations could enable LLMs to explicitly model and manipulate domain-specific symbolic rules (e.g., causal inference criteria, clinical constraints) within their reasoning process.

\paragraph{Representation Disentanglement.}
To enhance the compositionality and re-usability of latent concepts, dedicated representation learning methods aimed explicitly at disentanglement are necessary \cite{hsu2023disentanglement}. 
These methods would encourage latent dimensions or modules within neural architectures to correspond explicitly and uniquely to semantically meaningful, independently manipulable clinical concepts (e.g., drug–dose relations, disease–symptom clusters, risk–severity gradients) \cite{pandey2022disentangled}. Techniques such as variational inference with structured priors \cite{qin2023disentangled}, contrastive learning for conceptual differentiation \cite{hong2025latent}, and explicit regularization toward modularity \cite{wang2024disentangled, valentino2023multi, zhang2025learning} could help achieve more robust and interpretable disentangled representations.

\paragraph{Separating Reasoning from Ground Knowledge.}
Another critical direction involves the explicit architectural and functional separation of factual retrieval (ground knowledge) from reasoning mechanisms \cite{valentino2025mitigating}. 
Our experiments indicate that current models struggle when these functions are conflated. 
Developing modular architectures wherein factual knowledge bases (either latent or external) are distinct from inference engines could help clarify and strengthen the reasoning pipeline \cite{karpukhin2020dense}. This separation might involve training specialized neural retrieval components \cite{thai2024acr} paired with dedicated reasoning modules (e.g., transformer-based reasoning conditioned explicitly on retrieved symbolic or structured facts) \cite{vsevolodovna2025enhancing}. This approach can potentially improve transparency, debugging, and systematic enhancement of reasoning capabilities.

\paragraph{Explicit Counterfactual and Probabilistic Reasoning Modules.}
Given the observed failures in causal and probabilistic reasoning, future models could incorporate dedicated computational modules designed for counterfactual simulation and probabilistic abstraction \cite{ippolito2020causal, warren2024categorical, nguyen2024llms}. Such modules could explicitly represent and manipulate alternative hypothetical states (possible worlds), enabling robust reasoning about causal effects, clinical risks, and evidential coherence \cite{pearl2014probabilistic, wood2014new, liu2025large}. Techniques drawn from causal inference, Bayesian probabilistic programming, and differentiable inference frameworks \cite{carpenter2017stan, chater2006probabilistic, blei2017variational} would support the integration of rigorous statistical and causal reasoning within LLM architectures.

\paragraph{Benchmarks and Diagnostics for Compositional Reasoning.}
Future research should also focus on expanding and refining diagnostic benchmarks specifically targeting structured reasoning capabilities. This involves constructing large-scale, systematically controlled datasets where inferential correctness explicitly requires compositional integration of multiple structured facts \cite{sinha2019clutrr, fu2023seti, chen2024structtest}. Additionally, fine-grained diagnostic tools should be developed to pinpoint precise representational and inferential failures, thereby guiding targeted model improvements.

By pursuing these directions, we aim not merely to mitigate the surface-level failures observed in our study but to foster fundamentally more robust, interpretable, and generalizable models capable of reliable reasoning in high-stakes domains such as clinical decision-making.

\section{Related Work}
A dominant narrative in contemporary deep learning holds that enlarging model capacity and data exposure will monotonically improve not only task performance but also the \emph{quality} of internal representations. Foundational scaling-law studies \cite{kaplan2020scaling, hoffmann2022training} and flagship model reports \cite{brown2020language, touvron2023llama, bubeck2023sparks} implicitly endorse this view by demonstrating smooth performance gains as a function of compute and parameters. In contrast, \citet{marcus2022deep} and Others caution that competence on benchmarks need not imply the presence of robust, disentangled, and composable representations of the world. Most directly, \citet{kumar2025questioning} argue that scaling can yield increasingly \emph{entangled} latent spaces, rich in correlations yet brittle with respect to compositional, causal, or epistemic manipulation. Our work empirically instantiates this critique in the clinical domain: we show that models can store relevant medical facts while failing to \emph{deploy} them in structured inference.

A substantial body of evidence indicates that LLMs often exploit spurious artifacts rather than performing genuine reasoning. Early analyses uncovered annotation artifacts and “hypothesis-only” shortcuts in NLI datasets \cite{gururangan2018annotation, haraguchi2023discovering}; subsequent work has cataloged similar pattern-matching behaviour across tasks \cite{webson2022prompt}. In the LLM era, studies of CoT prompting reveal that the generated rationales can be unfaithful or post hoc, even when final answers are correct \cite{turpin2023language}. Relatedly, recent diagnostics locate “highly influential shortcuts” that guide model predictions independently of task semantics \cite{haraguchi2023discovering}. 

Prior efforts have investigated NLI in biomedical or clinical contexts \cite{jullien2023semeval, jullien2024semeval}, as well as targeted evaluations of LLMs on clinical trial reports or electronic health records \cite{rao2023assessing, liu2023utility}. These works primarily assess correctness on domain-specific entailment labels. Our contribution is complementary: we explicitly separate \emph{ground knowledge} from \emph{reasoning}. This design reveals a striking dissociation—near-ceiling factual recall alongside systematic inferential failure, thereby advancing the methodological landscape for evaluating clinical reasoning in LLMs.

\section{Conclusion}

Our evaluation reveals a consistent and domain-general limitation in the reasoning capabilities of current LLMs. Across four distinct integral clinical inference tasks, models repeatedly fail to perform structured, context-sensitive reasoning, even when they possess the relevant background knowledge. These failures are systematic, not stochastic, and reflect a deeper architectural deficit: the absence of explicit, coherent composable representations.

By formalizing the expected inference patterns for each task, we show that model behaviour diverges not only from clinical correctness but from principled reasoning itself. Rather than integrating facts compositionally or evaluating counterfactuals and evidential support, LLMs rely on shallow heuristics tied to surface form.

These findings suggest that scaling alone is insufficient. Reliable clinical reasoning, and robust reasoning more broadly, requires representational structure: a capacity to model, simulate, and reason over domain-grounded possibilities. Without it, the outputs of LLMs may remain fluent but fundamentally ungrounded.

\FloatBarrier
\bibliography{anthology,custom}
\bibliographystyle{acl_natbib}

\appendix

\section{Appendix}

\subsection{Limitations}

\paragraph{Benchmark scope and size.} Each reasoning family comprises ten instantiated items, generated via parametric templates. While this design enables tight control over targeted inferential properties, it limits statistical power and breadth. Our conclusions should be viewed as indicative rather than definitive.

\paragraph{Label granularity and evaluation.} We adopt a three-way NLI schema (entailment/contradiction/neutral). Clinical reasoning sometimes warrants graded or multi-label judgments (e.g., \emph{plausible but unsafe}, \emph{likely but not proven}). Our strict labels may under-represent partial reasoning success.

\paragraph{Prompting and sampling choices.} Only two prompt formats (Direct vs.\ CoT) and ten stochastic completions were considered. Different temperatures, self-consistency decoding, tool augmentation, or verifier-guided prompting might alter outcomes.

\paragraph{Model coverage and opacity.} The selection, while diverse, is not exhaustive. Proprietary models are black boxes: training data, alignment procedures, and internal tooling are undisclosed, constraining interpretability. Results may not generalize to future model versions.

\paragraph{Ethical and deployment considerations.} Demonstrating LLM failures in high-stakes medicine underscores risk but does not specify safe deployment guidelines. Our benchmark should complement, not replace, rigorous validation, monitoring, and regulatory oversight.

\subsection{Prompt Choice Effects}
Table~\ref{tab:prompt_delta} compares \textit{Direct} versus \textit{Chain-of-Thought (CoT)} prompting across all four reasoning families and six models. CoT yielded a small but consistent absolute gain in mean accuracy (+0.03 overall; 0.226 vs.\ 0.196), yet this effect was neither monotonic across models nor uniform across tasks. For instance, GPT-4o showed no aggregate change (0.15$\rightarrow$0.15), while GPT-4o-mini and LLaMA~3.2 benefited most (+0.078 and +0.058, respectively). Conversely, OpenAI o3 slightly regressed ($-0.005$). These heterogeneous deltas, combined with the limited item count (40 main-task instances), caution against strong causal claims about CoT’s efficacy. 

Qualitative inspection of CoT traces suggests that additional verbalization often \emph{explains} a pre-selected answer rather than steering inference: many rationales restate lexical cues (e.g., numerical percentages or clinician assertions) without invoking the formal criteria (counterfactual comparison, constraint checking, probabilistic harm). In short, CoT increases verbosity but does not reliably mitigate heuristic drift. Future work should dissociate “reasoning exposure” from “reasoning control”.

\begin{table}[h!]
\centering
\small
\begin{tabular}{lccc}
\toprule
\textbf{Model} & \textbf{CoT} & \textbf{Direct} & $\Delta$ \\
\midrule
DeepSeek R1   & 0.283 & 0.233 & +0.050 \\
Gemini 2.5    & 0.285 & 0.285 & +0.000 \\
OpenAI o3     & 0.223 & 0.228 & $-$0.005 \\
GPT-4o        & 0.150 & 0.150 & +0.000 \\
GPT-4o-mini   & 0.183 & 0.105 & +0.078 \\
LLaMA 3.2 3B  & 0.233 & 0.175 & +0.058 \\
\midrule
\textbf{Overall} & \textbf{0.226} & \textbf{0.196} & \textbf{+0.030} \\
\bottomrule
\end{tabular}
\caption{Mean main-task accuracy across four tasks for each model under CoT vs.\ Direct prompting.}
\label{tab:prompt_delta}
\end{table}

\subsection{Model Choice Effects}
Model identity explained more variance than prompt format. Gemini~2.5 and o3 dominated Causal Attribution (0.73–0.77), yet both collapsed on Compositional Grounding (0.00–0.02). DeepSeek R1, while never best overall, was uniquely strongest on Compositional Grounding (0.19 CoT, 0.14 Direct) and competitive on Epistemic Verification (0.31–0.32). LLaMA~3.2, despite markedly lower ground-knowledge scores, topped Risk Abstraction (0.27 CoT; 0.25 Direct) and Epistemic Verification under Direct prompting (0.33). This pattern indicates that absolute parameter count is not a reliable proxy for structured reasoning competence: lighter open models sometimes matched or surpassed flagship models on specific inference types.

Two further observations emerge. First, ceiling-level GKMRV accuracy across nearly all proprietary models suggests comparable factual coverage; the divergence appears in \emph{deployment}, not \emph{possession}, of knowledge. Second, inter-task rank orderings were weakly correlated: excelling in causal attribution did not predict strength in compositional or epistemic reasoning. This orthogonality motivates task-specific diagnostics rather than omnibus “reasoning scores,” and underscores the need for architectural or training interventions targeted at particular inferential primitives (e.g., constraint satisfaction, counterfactual simulation).

\subsection{Prompts}
\begin{table}[h!]
\centering
\small
\begin{tabular}{>{\ttfamily\raggedright\arraybackslash}p{0.9\columnwidth}}
\toprule
\textbf{Direct Prompt} \\
\midrule
You are given a premise and a statement. Your task is to determine the relationship between the statement and the premise. You must end your Answer with \texttt{output: entailment}, \texttt{output: neutral}, or \texttt{output: contradiction}. \\[0.8ex]
\textbf{statement}: \texttt{\{statement\}} \\
\textbf{premise}: \texttt{\{premise\}} \\
\textbf{Answer}: \\
\bottomrule
\end{tabular}
\caption{Direct prompt.}
\label{tab:prompt_nli}
\end{table}

\begin{table}[h!]
\centering
\small
\begin{tabular}{>{\ttfamily\raggedright\arraybackslash}p{0.9\columnwidth}}
\toprule
\textbf{CoT Prompt} \\
\midrule
You are given a premise and a statement. Your task is to determine the relationship between the statement and the premise. Explain your reasoning step by step. You must end your Answer with \texttt{output: entailment}, \texttt{output: neutral}, or \texttt{output: contradiction}. \\[0.8ex]
\textbf{statement}: \texttt{\{statement\}} \\
\textbf{premise}: \texttt{\{premise\}} \\
\textbf{Answer}: \\
\bottomrule
\end{tabular}
\caption{CoT prompt.}
\label{tab:prompt_cot}
\end{table}

\begin{table}[h!]
\centering
\small
\begin{tabular}{>{\ttfamily\raggedright\arraybackslash}p{0.9\columnwidth}}
\toprule
\textbf{Direct Prompt} \\
\midrule
You are given a premise and a statement.
Your task is to determine whether the statement is factually correct based on the clinical information in the premise and established medical knowledge.
You must end your response with output: True or output: False\\[0.8ex]
\textbf{statement}: \texttt{\{statement\}} \\
\textbf{premise}: \texttt{\{premise\}} \\
\textbf{Answer}: \\
\bottomrule
\end{tabular}
\caption{Direct GKMRV prompt.}
\label{tab:prompt_GKMRV}
\end{table}

\begin{table}[h!]
\centering
\small
\begin{tabular}{>{\ttfamily\raggedright\arraybackslash}p{0.9\columnwidth}}
\toprule
\textbf{CoT Prompt} \\
\midrule
You are given a premise and a statement.
Your task is to determine whether the statement is factually correct based on the clinical information in the premise and established medical knowledge.
Explain your reasoning step by step.
You must end your response with output: True or output: False \\[0.8ex]
\textbf{statement}: \texttt{\{statement\}} \\
\textbf{premise}: \texttt{\{premise\}} \\
\textbf{Answer}: \\
\bottomrule
\end{tabular}
\caption{CoT GKMRV prompt.}
\label{tab:prompt_cot_GKMRV}
\end{table}

\end{document}